\documentclass{article} % For LaTeX2e
\usepackage{iclr2025_conference,times}

\usepackage{amsmath,amsfonts,bm}

\newcommand{\captiona}{{\em (a)}}
\newcommand{\captionb}{{\em (b)}}
\newcommand{\captionc}{{\em (c)}}

\def\eqref#1{equation~\ref{#1}}
\def\1{\bm{1}}

\DeclareMathAlphabet{\mathsfit}{\encodingdefault}{\sfdefault}{m}{sl}
\SetMathAlphabet{\mathsfit}{bold}{\encodingdefault}{\sfdefault}{bx}{n}

\newcommand{\R}{\mathbb{R}}

\DeclareMathOperator{\Tr}{Tr}

\newcommand{\bu}{\boldsymbol{u}}

\newcommand{\bx}{\boldsymbol{x}}

\newcommand{\bA}{\boldsymbol{A}}

\newcommand{\bH}{\boldsymbol{H}}

\newcommand{\btheta}{\boldsymbol{\theta}}

\newcommand{\cD}{\mathcal{D}}

\newcommand{\cL}{\mathcal{L}}

\newcommand{\norm}[1]{\left\lVert#1\right\rVert}

\usepackage{url}
\usepackage{xcolor}
\usepackage{amsmath}
\usepackage{graphicx}
\usepackage[colorlinks=true, allcolors=blue]{hyperref}
\usepackage{times}  % DO NOT CHANGE THIS
\usepackage{helvet}  % DO NOT CHANGE THIS
\usepackage{courier}  % DO NOT CHANGE THIS
\usepackage{natbib}  % DO NOT CHANGE THIS AND DO NOT ADD ANY OPTIONS TO IT
\usepackage{caption} % DO NOT CHANGE THIS AND DO NOT ADD ANY OPTIONS TO IT
\usepackage{algorithm}
\usepackage{booktabs}
\usepackage{newfloat}
\usepackage{listings}
\usepackage{mathtools}
\usepackage{multirow}
\usepackage{bbm}
\usepackage{amsfonts,amssymb}
\usepackage{graphicx}
\usepackage{wrapfig}

\usepackage{amsmath}
\usepackage{amssymb}
\usepackage{mathtools}
\usepackage{amsthm}

\usepackage[shortlabels]{enumitem} % For itermization
\setlist[itemize]{topsep=0em, itemsep=0em, partopsep=0em, parsep=0.5em}

\allowdisplaybreaks
\usepackage{mdframed}
\usepackage[normalem]{ulem}
\usepackage[most]{tcolorbox}  

\usepackage[capitalize,noabbrev,nameinlink]{cleveref}[0.19]

\definecolor{darkgrey}{rgb}{0.53,0.53,0.53}
\definecolor{middlegrey}{rgb}{0.53,0.53,0.53}
\definecolor{mygrey}{rgb}{0.9,0.9,0.9}
\definecolor{mydarkblue}{rgb}{0,0.08,0.45}
\definecolor{darkdarkblue}{rgb}{0.0,0.0,0.3}
\definecolor{darkblue}{rgb}{0.0,0.0,0.7}
\definecolor{darkred}{rgb}{0.4,0,0.3}
\definecolor{lightblue}{HTML}{EDE6DC}
\definecolor{lightred}{HTML}{FFFAFA}
\definecolor{fancyblue}{HTML}{4771E3}
\definecolor{grey}{rgb}{0.95,0.95,0.95}

\hypersetup{
    colorlinks=true,       
    linkcolor=darkred,        
    citecolor=darkblue,  
    filecolor=darkblue,
    urlcolor=darkblue       
}

\theoremstyle{plain}

\theoremstyle{definition}

\theoremstyle{remark}

\title{On the Cone Effect in the Learning Dynamics}

\author{%
Zhanpeng Zhou\textsuperscript{1}$^{\star}$,
Yongyi Yang\textsuperscript{2}, 
Jie Ren\textsuperscript{1},
Mahito Sugiyama\textsuperscript{3,4},
Junchi Yan\textsuperscript{1}\thanks{Corresponding author.}\\
\textsuperscript{1}Sch. of Computer Science \& Sch. of Artificial Intelligence, Shanghai Jiao Tong University \\
\textsuperscript{2}University of Michigan
\textsuperscript{3}National Institute of Informatics
\textsuperscript{4}The Graduate University for\\
Advanced Studies, SOKENDAI\\
\texttt{\small \{zzp1012,yanjunchi\}@sjtu.edu.cn}\\
}

\iclrfinalcopy % Uncomment for camera-ready version, but NOT for submission.
\begin{document}

\maketitle

\begin{abstract}
Understanding the learning dynamics of neural networks is a central topic in the deep learning community. 
In this paper, we take an empirical perspective to study the learning dynamics of neural networks in real-world settings. 
Specifically, we investigate the evolution process of the empirical Neural Tangent Kernel (eNTK) during training.
Our key findings reveal a two-phase learning process: i) in Phase I, the eNTK evolves significantly, signaling the rich regime, and ii) in Phase II, the eNTK keeps evolving but is constrained in a narrow space, a phenomenon we term the cone effect. 
This two-phase framework builds on the hypothesis proposed by \citet{fort2020deep}, but we uniquely identify the cone effect in Phase II, demonstrating its significant performance advantages over fully linearized training.
\end{abstract}

\vspace{-7pt}
\section{Introduction}
\vspace{-7pt}
Research on the learning dynamics of neural networks has drawn considerable attention in the learning theory community.
For example, \citet{jacot2018neural,Chizat2019lazytraining,yang2021tp4,arora2019on} studied the infinitely wide neural networks, where the dynamics are linear and can be captured by a static kernel function, commonly referred to as the Neural Tangent Kernel (NTK).
This training behavior is known as the \emph{lazy regime}.
However, the lazy regime deviates significantly from real-world scenarios and cannot account for the surprising generalization ability of neural networks~\citep{Chizat2019lazytraining}.
A series of recent studies aim to uncover the complex non-linear learning dynamics beyond the lazy regime, namely the \emph{rich regime}~\citep{geiger2020disentangling,Woodworth2020kernel}, but they generally focus on extremely simplified models.
Studying the learning dynamics in practice remains an area of open research.

\textbf{Our contributions.}
In this work, we investigate the learning dynamics of neural networks in real-world scenarios.
Specifically, we investigate the evolution process of the empirical Neural Tangent Kernel (eNTK) during training.
More precisely, we find that: \begin{itemize}[leftmargin=0.30in]
    \item In \textbf{Phase I}, the eNTK evolves significantly. Neural networks exhibit highly non-linear dynamics, signaling the rich regime.
    \item In \textbf{Phase II}, the eNTK keeps evolving but are constrained in a narrow space.
    This training behavior differs from the lazy regime significantly, namely the \emph{cone effect}.
\end{itemize}
Our investigation follows \citet{fort2020deep}, where they proposed the two-phase hypothesis (see \cref{sec:prelim}) to explain the Linear Mode Connectivity phenomenon~\citep{frankle2020linear,zhou2023LLFC}.
Similarly, they noted the two-phase behavior in the learning dynamics; however, they considered the eNTK to be approximately fixed in Phase II, i.e., the lazy regime.
Surprisingly, other than the lazy regime, we discover the cone effect phenomenon in the second phase, and further show that this cone effect yields non-negligible performance benefits over the lazy regime.

\vspace{-7pt}
\section{Related Work}
\label{sec:related}
\vspace{-7pt}

\textbf{Lazy Regimes.}
Numerous theoretical studies~\citep{du2018gradient,li2018learning,du2019gradient,allenzhu2019convergence,zou2020gradient} proved that over-parameterized models can achieve zero training loss with minimal parameter variation.
Moreover, \citet{jacot2018NTK,yang2019scaling,arora2019on,lee2019wide} showed that the learning dynamics of infinitely wide neural networks can be captured by a frozen kernel at initialization, known as the Neural Tangent Kernel (NTK).
This behavior, often termed as \emph{lazy regime}, typically occurs in over-parameterized models with large initialization and is considered undesirable in practice~\cite{Chizat2019lazytraining}.

\textbf{Rich Regimes.}
In contrast to the lazy regime, where learning dynamics are linear, the rich regime, or feature learning, exhibits complex nonlinear dynamics~\citep{geiger2020disentangling,jacot2021saddle,xu2024does}. 
Previous work has established that the transition between these regimes depends critically on initialization parameters.
For instance, \citet{geiger2020disentangling,Woodworth2020kernel} showed that the absolute scale of network initializations governs this transition.
Subsequent work further revealed that the relative scale of initializations~\citep{azulay2021shape,kunin2024get} and their effective rank~\citep{liu2024how} can similarly induce feature learning.

\vspace{-7pt}
\section{Background and Preliminaries}
\label{sec:prelim}
\vspace{-7pt}

\textbf{Notation and Setup.}
In this work, we focus on a classification task. 
Denote $[k]:= \{1, 2, \cdots, k\}$.
Let $\cD=\{(\bx_i,y_i)\}_{i=1}^n$ be the training set of size $n$, where $\bx_i \in \R^{d_0}$ represents the $i$-th input and $y_i\in [c]$ represents the corresponding target.
Here, $c$ is the number of classes.
Let $f: \cD \times \R^p \to \R$ be the neural network.
Then $f(\bx, \btheta) \in \R$ denotes the output of model $f$ on the input $\bx$, where $\btheta \in \R^p$ represents the model parameters.
Let $\ell(f(\bx_i, \btheta), y_i)$ be the loss at the $i$-th data point, simplified to $\ell_i(\btheta)$.
The total loss over the dataset $\cD$ is then denoted as $\cL_{\cD}(\btheta) = \frac{1}{n}\sum_{i=1}^n \ell_i(\btheta)$.
Additionally, we use bold lowercase letters (e.g., $\bx$) to denote vectors, and bold uppercase letters (e.g., $\bA$) to represent matrices. 
For a matrix $\bA$, let $\norm{\bA}_2$, $\norm{\bA}_F$, and $\Tr(A)$ denote its spectral norm, Frobenius norm, and trace, respectively.

\textbf{Training Dynamics, Neural Tangent Kernel.}
Consider minimizing the total loss $\cL_{\cD}(\btheta)$ using gradient flow; the update rule can be written as follows:
\begin{align}
    \frac{d \btheta(t)}{d t} = -\nabla_{\btheta} \cL (\btheta(t)) = -\sum_{i=1}^n \left(\frac{\partial f_t(\bx_i)}{\partial \btheta}\right)^{\top} \frac{\partial \cL_{\cD}(\btheta(t))}{\partial f_t(\bx_i)},
\end{align}
where we denote $f_t(\bx_i):=f(\bx_i, \btheta(t))$ for simplicity.
Subsequently, the evolution of the network output $f_t(\bx_i)$ can be written as follows.
\begin{equation}
    \frac{d f_t(\bx_i)}{d t} = -\sum_{j=1}^n \left\langle \frac{\partial f_t(\bx_i)}{\partial \btheta}, \frac{\partial f_t(\bx_j)}{\partial \btheta} \right\rangle \frac{\partial \cL(\btheta(t))}{\partial f_t(\bx_i)}. \label{eq:func_learn_single}
\end{equation}
Denoting $\bu(t)=\{f_t(\boldsymbol{x}_i)\}_{i=1}^n \in \R^n$ as the network outputs for all inputs, then a more compact form of \cref{eq:func_learn_single} is given by:
\begin{equation}
\label{eq: eNTK}
\frac{d \bu(t)}{d t}= - \bH(\btheta(t)) \nabla_{\bu(t)}\cL(\btheta(t)),
\end{equation}
where $\boldsymbol{H}(\btheta(t)) \in \R^{n\times n}$ is a kernel matrix and its $(i,j)$-th entry is {\small $\left\langle \frac{\partial f_t(\bx_i)}{\partial \btheta}, \frac{\partial f_t(\bx_j)}{\partial \btheta} \right\rangle$}. 
This matrix is referred to as the \emph{empirical neural tangent kernel (eNTK)}.
For an infinitely wide neural network, the matrix $\bH(\btheta(t)) \approx \bH(\btheta(0))$ remains nearly constant throughout training~\citep{arora2019exact,jacot2018neural}, which is also known as the \textbf{lazy regime}~\citep{domine2024lazy}.

In comparison, for neural networks with finite width, the eNTK matrix $\boldsymbol{H}(\btheta(t))$ evolves significantly during training, commonly termed as the \textbf{rich regime}.
To quantify the changes in the eNTK matrix during training, \citet{fort2020deep} introduced two metrics: \emph{kernel distance} and \emph{kernel velocity}.
The kernel distance $S(\btheta, \btheta^\prime)$  measures the difference between the kernel matrices at two different points, i.e., $\btheta$ and $\btheta^\prime$, in a scale-invariant way:
\begin{equation}\label{eq: kernel distance}
    S(\btheta, \btheta^\prime) \triangleq 1 - \frac{\Tr(\boldsymbol{H}(\btheta)\boldsymbol{H}^{\top}(\btheta^\prime))}{\sqrt{\Tr(\boldsymbol{H}(\btheta)\boldsymbol{H}^{\top}(\btheta))} \sqrt{\Tr(\boldsymbol{H}(\btheta^\prime)\boldsymbol{H}^{\top}(\btheta^\prime))}}
\end{equation}
Building on the kernel distance, the kernel velocity $v(t)$ captures the rate at which the kernel changes at a given iteration $t$:
\begin{equation}
\label{eq: kernel velocity}
    v(t) \triangleq \frac{S(\btheta(t),\btheta(t+d t))}{d t}
\end{equation}
A large kernel velocity indicates that the kernel is evolving rapidly during training, which is characteristic of the rich regime. 
Together, these metrics provide a framework for analyzing the dynamics of neural network training, offering implications for learning behavior.

\textbf{The Two-Phase Hypothesis.}
\citet{fort2020deep} discovered a two-phase phenomenon in the training dynamics of neural networks via the \emph{kernel velocity}.
Specifically, they found that the kernel velocity is relatively large during the first few training epochs but drops rapidly after that.
They also showed that using standard training followed by linearized training performs similarly to using only standard training with a small learning rate.
Based on these observations, they proposed the following two-phase hypothesis:

\begin{tcolorbox}[notitle, rounded corners, colframe=middlegrey, colback=lightblue, 
       boxrule=1.5pt, boxsep=0pt, left=0.15cm, right=0.17cm, enhanced, 
       toprule=1.5pt,
    ]
\begin{itemize}[leftmargin=0.30in]
    \item \textit{\fontsize{9.5pt}{10pt}\selectfont In \textbf{Phase I}, neural networks exhibit highly non-linear dynamics, signaling the rich regime.}
    \item \textit{\fontsize{9.5pt}{10pt}\selectfont In \textbf{Phase II}, the dynamics are approximately linear, and the models enter the lazy regime.}
\end{itemize}

\end{tcolorbox}

\textbf{Main Experimental Setup.}
Following \citet{frankle2020linear}, we perform our experiments on commonly used image classification datasets MNIST \citep{lecun1998gradient} and CIFAR-10~\citep{krizhevsky2009learning}, and with the standard network architectures ResNet-20~\citep{kaiming2016residual}, VGG-16~\citep{simonyan2015vgg} and LeNet~\citep{lecun1998gradient}.  
We follow the same training procedures and hyperparameters as in \citet{frankle2020linear,ainsworth2023git}.

\vspace{-7pt}
\section{The Cone Effect in Learning Dynamics}
\label{sec:cone}
\vspace{-7pt}

Previous studies empirically characterized the training dynamics of neural networks into two phases, transitioning from rich to lazy regimes.
In this section, we dig deeper into this picture and observe that in the second phase, rather than the lazy regime, the eNTK matrices keep evolving but in a constrained space, namely \emph{the cone effect}.

\begin{figure}[tb!]
    \centering
    \includegraphics[width=0.94\linewidth]{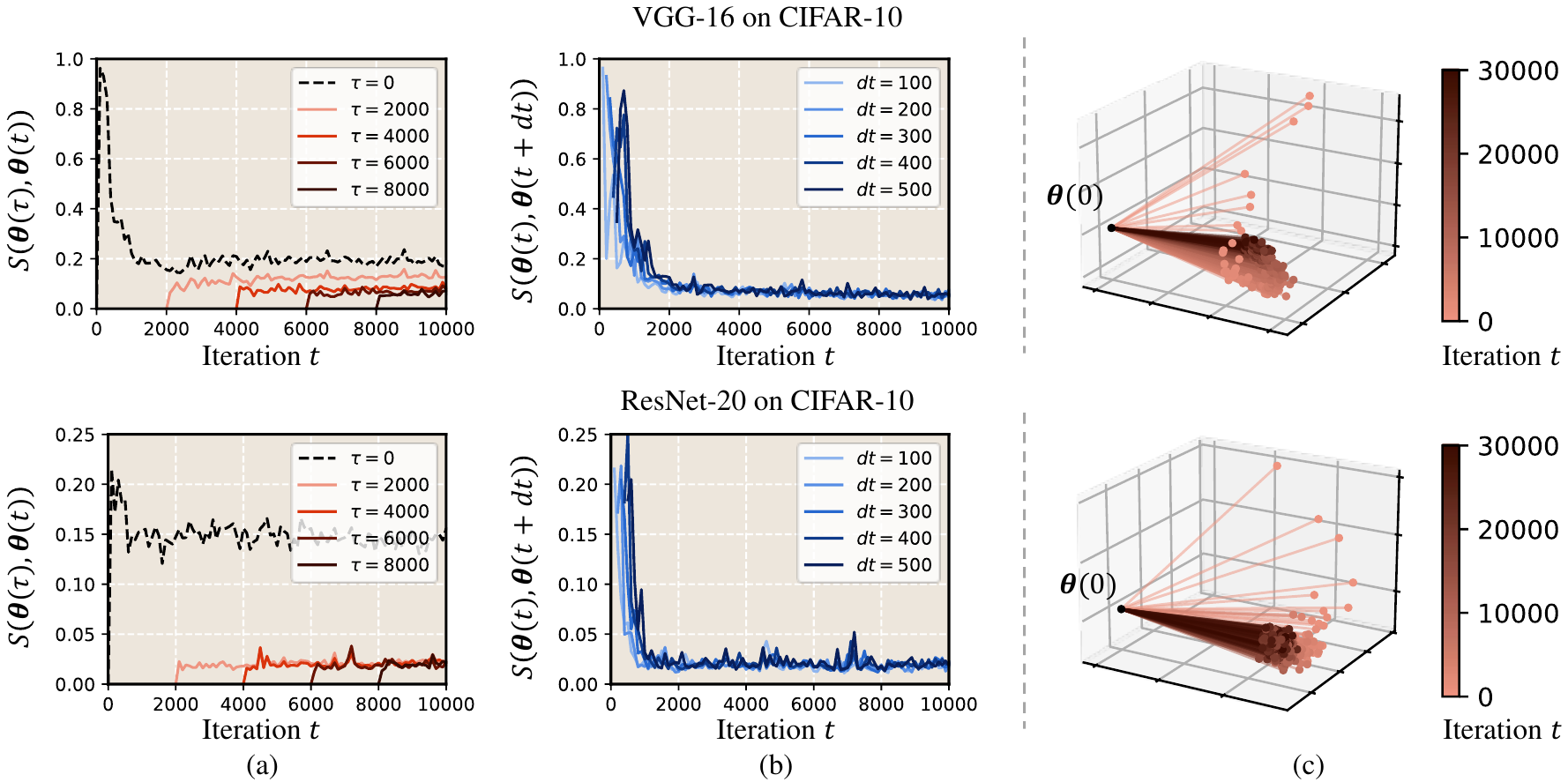}
    \caption{\textbf{The cone effect in the learning dynamics.}
    \textbf{(a)} 
    The kernel distance between the current iterate $\btheta(t)$ and a reference point $\btheta(\tau)$ v.s. training iteration $t$, where $\tau$ is varied.
    \textbf{(b)}
    The kernel distance between two adjacent iterates $\btheta(t)$ and $\btheta(t+dt)$ vs. training iteration $t$, where $dt$ is varied.
    \textbf{(c)}
    The visualization of the changes of the eNTK matrices $\bH(t)$.
    The black dot represents position of the eNTK matrix at initialization, i.e., $\bH(0)$.
    The other dot represents the relative position of $\bH(t)$ at $t>0$, with darker color indicating larger iteration.
    }
    \label{fig: cone_effect}
    \vspace{-10pt}
\end{figure}

\textbf{Going Beyond the Lazy Regime: The Cone Effect.}
Through the kernel distance, we delve into the two-phase hypothesis in the learning dynamics.
First, we compute the kernel distance between the eNTKs at two adjacent points $\btheta(t)$ and $\btheta(t+dt)$.
In \cref{fig: cone_effect}~\captionb, we observe that, across different values of $dt$, the adjacent kernel distance $S(\btheta(t), \btheta(t+dt))$ is significant in the early training phase and then drops quickly to a low but non-negligible value.
This result aligns with the two-phase hypothesis, where in Phase I (approximately $0<t<2000$) the model learns in the rich regime and the eNTK evolves significantly; in Phase II, the model is in the lazy regime and the eNTK is approximately fixed.
However, surprisingly, we note that in Phase II, the values of $S(\btheta(t), \btheta(t+dt))$ are upper-bounded by the same value for different $dt$.
One possible explanation for this phenomenon is that in Phase II, the eNTK evolves in a constrained space.
To validate this, we further measure how the distance between the eNTK matrices at the current iterate $\btheta(t)$ and a referent point $\btheta(\tau)$ changes during training.
As shown in \cref{fig: cone_effect}~\captiona, for $\tau\in \{2000, 4000, 6000, 8000\}$, 
the kernel distance between the current iterate and the reference point, i.e., $S(\btheta(t), \btheta(\tau))$, first increases and then keeps nearly constant during training.
This result verifies our claim and suggests that in Phase II, beyond the lazy regime, the model learns in a constrained function space.
The visualization in \cref{fig: cone_effect}~\captionc~ further confirms our picture, where a clear ``cone'' pattern is observed.

\begin{figure}
    \centering
    \includegraphics[width=0.65\linewidth]{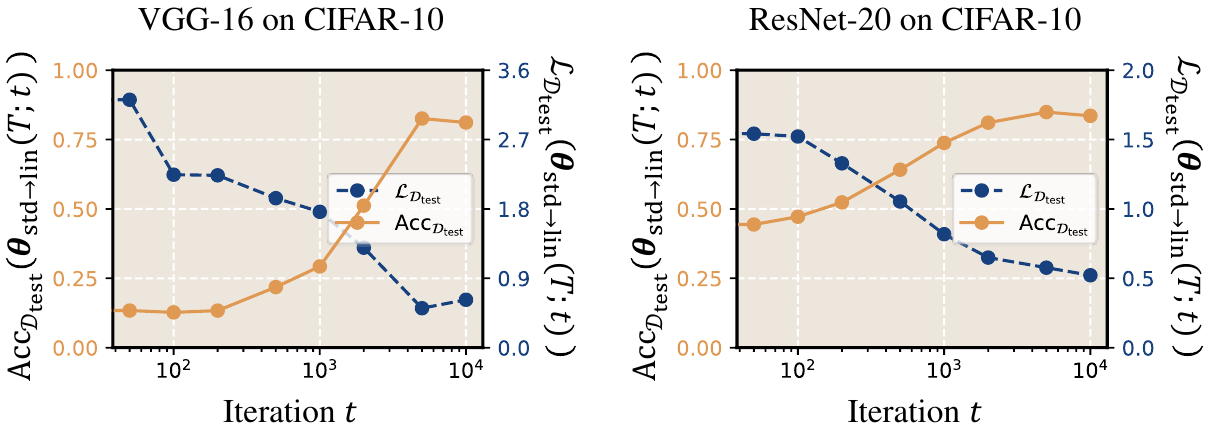}
    \caption{\textbf{Non-linear advantages of the cone effect.}
    Test accuracy $\text{Acc}_{\cD_{\rm test}}(\btheta_{\rm std \to lin}(T; t))$ \textbf{(left)} and test loss $\cL_{\cD_{\rm test}}(\btheta_{\rm std \to lin}(T; t))$ vs. the switching iteration $t$.
    $\btheta_{\rm std \to lin}(T; t)$ denotes the model initially trained with standard method up to iteration $t$, followed by linearized training up to iteration $T$. 
    $T$ is set to $10^{4}$.
    }
    \label{fig: advantage}
    \vspace{-10pt}
\end{figure}

\textbf{Non-Linear Advantages of the Cone Effect.}
Though in Phase II the learning happens in a constrained function space, it still provides significant advantages over the completely lazy regime.
To investigate this, we consider a ``switching'' training method: we first train a neural network with a standardized training method, and switch to the linearized training (corresponding to the completely lazy regime) until $T$ iterations.
We vary the switching point $t$ and obtain different solutions $\btheta_{\rm std \to lin}(T; t)$.
In \cref{fig: advantage}, we observe that the test performance of $\btheta_{\rm std \to lin}(T; t)$ generally increases with $t$, especially when $t > 2000$. 
This result implies that the cone effect in Phase II still offers significant advantages over the solely lazy regime.

Together, we conclude the following conjecture:
\begin{tcolorbox}[notitle, rounded corners, colframe=middlegrey, colback=lightblue, 
       boxrule=1.5pt, boxsep=0pt, left=0.15cm, right=0.17cm, enhanced, 
       toprule=1.5pt,
    ]
\textbf{The Cone Effect}: \textit{\fontsize{9pt}{10pt}\selectfont In \textbf{Phase II}, beyond the lazy regime, the model learns in a constrained function space and the performance improves significantly.}
\end{tcolorbox}

\begin{figure}[tb!]
    \centering
    \includegraphics[width=0.94\linewidth]{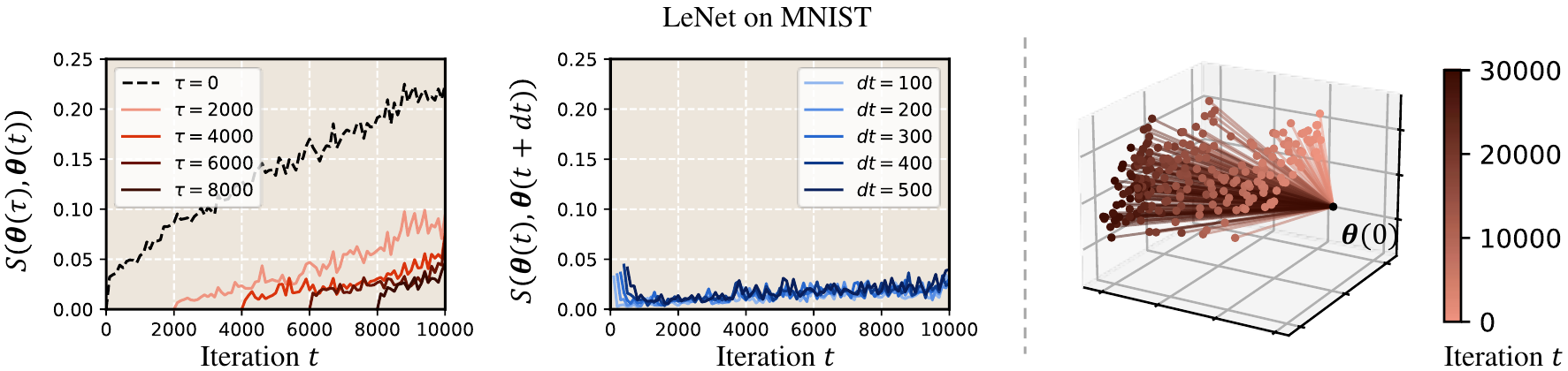}
    \caption{\textbf{A counterexample for the two-phase hypothesis.}
    \textbf{(a)} 
    The kernel distance between the current iterate $\btheta(t)$ and a reference point $\btheta(\tau)$ v.s. training iteration $t$, where $\tau$ is varied.
    \textbf{(b)}
    The kernel distance between two adjacent iterates $\btheta(t)$ and $\btheta(t+dt)$ vs. training iteration $t$, where $dt$ is varied.
    \textbf{(c)}
    The visualization of the changes of the eNTK matrices $\bH(t)$.
    The black dot represents position of the eNTK matrix at initialization, i.e., $\bH(0)$.
    The other dot represents the relative position of $\bH(t)$ at $t>0$, with darker color indicating larger iteration.
    }
    \label{fig: counterexample}
    \vspace{-10pt}
\end{figure}

\vspace{-7pt}
\section{Conclusion and Limitations}
\vspace{-7pt}
In summary, our work unveils a cone effect in the late learning dynamics. 
However, the cone effect and the two-phase hypothesis are not universal.
In \cref{fig: counterexample}, we observe a counterexample.
Specifically, the kernel keeps evolving during training, and no clear two-phase pattern is observed.
Further efforts on uncovering the contributing factors of the cone effect are expected.

\bibliography{iclr2025_conference}
\bibliographystyle{iclr2025_conference}

\end{document}